\documentclass[10pt,twocolumn,letterpaper]{article}

\usepackage[pagenumbers]{cvpr}     

\usepackage{algorithm}
\usepackage{algpseudocode}
\usepackage{multirow}
\usepackage{adjustbox}

\usepackage{tabularx} 
\newcommand{\modelname}{SplatFill}

\usepackage[dvipsnames]{xcolor}

\definecolor{cvprblue}{rgb}{0.21,0.49,0.74}
\usepackage[pagebackref,breaklinks,colorlinks,citecolor=cvprblue]{hyperref}

\def\paperID{206} 
\def\confName{3DV\xspace}
\def\confYear{2026\xspace}

\title{\modelname{}: 3D Scene Inpainting via Depth-Guided Gaussian Splatting}

\author{Mahtab Dahaghin \and Milind G. Padalkar \and Matteo Toso \and Alessio Del Bue\\
Pattern Analysis and Computer Vision (PAVIS)\\
Istituto Italiano di Tecnologia (IIT)\\
{\tt\small \{mahtab.dahaghin, milind.padalkar, matteo.toso, alessio.delbue\}@iit.it}
}

\begin{document}


\maketitle


\begin{abstract}

3D Gaussian Splatting (3DGS) has enabled the creation of highly realistic 3D scene representations from sets of multi-view images. However, inpainting missing regions, whether due to occlusion or scene editing, remains a challenging task, often leading to blurry details, artifacts, and inconsistent geometry. In this work, we introduce SplatFill, a novel depth-guided approach for 3DGS scene inpainting that achieves state-of-the-art perceptual quality and improved efficiency. Our method combines two key ideas: (1) joint depth-based and object-based supervision to ensure inpainted Gaussians are accurately placed in 3D space and aligned with surrounding geometry, and (2) we propose a consistency-aware refinement scheme that selectively identifies and corrects inconsistent regions without disrupting the rest of the scene. Evaluations on the SPIn-NeRF dataset demonstrate that SplatFill not only surpasses existing NeRF-based and 3DGS-based inpainting methods in visual fidelity but also reduces training time by 24.5\%. Qualitative results show our method delivers sharper details, fewer artifacts, and greater coherence across challenging viewpoints.

\end{abstract}

\section{Introduction}
\label{sec:intro}

\begin{figure}
  \centering
  \includegraphics[width=0.5\textwidth]{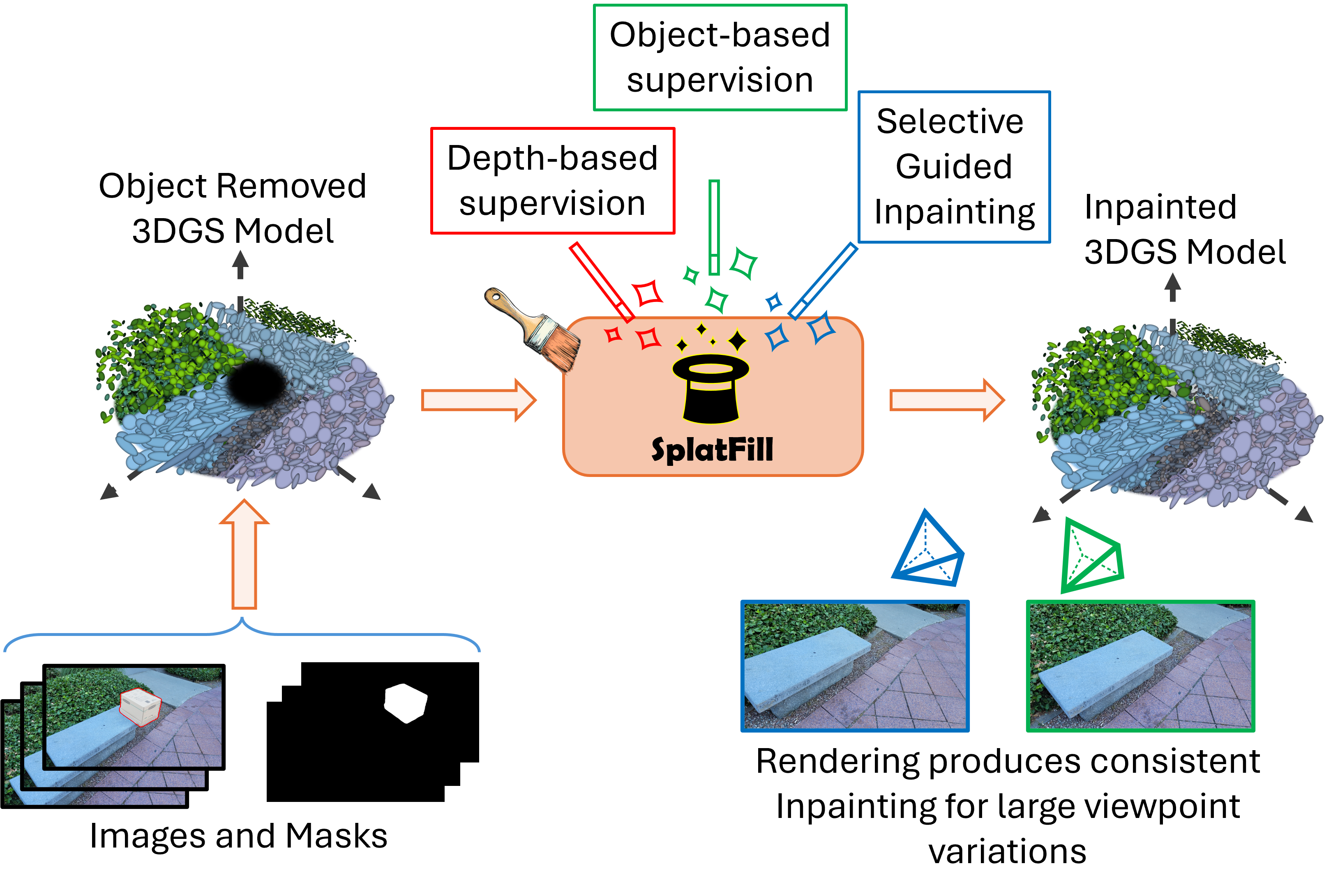}
  \caption{The \textit{3D scene inpainting} problem: 
  Starting from a set of images with known poses and masks occluding the area to be inpainted, we train a model to incrementally fill the masked regions with high-frequency geometric and texture details that remain consistent with the unmasked scene elements. The resulting method is consistent even under strong viewpoint changes.
  } 
  \label{fig:intro}
\end{figure}

In recent years, Novel View Synthesis (NVS) methods such as NeRF~\cite{mildenhall2020nerf} and 3DGS~\cite{Kerbl20233dgs} have made it possible to fully capture the appearance of a scene starting from a small set of images, training a model to generate photorealistic, accurate renderings from any viewpoint. Such methods have the potential to play a significant role in applications such as movie editing, game design, virtual/augmented reality, and robotics. However, in real-world applications, static reconstruction is not enough. To fully exploit NVS within these fields, we would require complete control over the 3D environment, \ie being able to edit the scene dynamically by removing, replacing and modifying objects in the model. Such changes would expose previously unseen surfaces, which were either occluded by other elements or in contact with the edited components, resulting in visible gaps in the 3D model. The task of completing these missing elements, outlined in Fig.~\ref{fig:intro}, is commonly referred to as \emph{3D scene inpainting}, and requires high-quality visual synthesis as well as geometric consistency across all viewpoints. These requirements are hard to satisfy, making 3D scene inpainting a still open challenge.

Inpainting missing content is a long-standing problem in computer vision. Image inpainting methods, from early PDE- and patch-based approaches~\cite{bertalmio2000image, criminisi2004inpainting, barnes2009patchmatch, herling2012pixmix} to modern diffusion-based generative models~\cite{rombach2022stablediffusion,Lugmayr2022repaint,xie2023smartbrush,Corneanu2024latentinpaint}, have achieved impressive results in 2D. Yet, simply applying image inpainting independently to each reference view in a multi-view set is insufficient: minor inconsistencies between views quickly accumulate into artifacts and blurry results when rendered in 3D. This multi-view coherence requirement is the primary obstacle separating 3D scene inpainting from its 2D counterpart.

Recent works have explored 3D scene inpainting for both NeRF~\cite{mirzaei2023spin, mirzaei2023reference, wang2024innerf360, weber2023nerfiller, MVIPNeRF, lin2024maldnerf, yin2023or, prabhu2023inpaint3d} and 3DGS~\cite{liu2024infusion, mirzaei2024reffusion, wang2025gscream}. While NeRF-based methods generally achieve high-quality results, their slow training and rendering speeds limit real-time applications. Conversely, 3DGS-based methods offer significant speed advantages but still struggle to achieve high-quality, consistent inpainting across all views. Existing inpainting approaches typically follow one of two main paradigms. 
The first conditions the reconstruction on a \emph{single} inpainted reference view~\cite{wang2025gscream}, which can yield accurate renderings from viewpoints close to that reference but tends to fail under large viewpoint changes. 
The second aggregates inpainted 2D images from \emph{multiple} viewpoints~\cite{cao2024mvinpainter, salimi2025geometry} in an attempt to enforce multi-view consistency. 
However, because the generative inpainting process does not always produce perfectly consistent 2D inpainting results across different views, enforcing 3D consistency can introduce artifacts and lead to blurring or unrealistic renderings.

We introduce \textit{\modelname{}}, a new 3DGS-based inpainting framework that overcomes these limitations by integrating high-detail single-view inpainting with a multi-view refinement process guided by geometric priors. Our method captures fine details from a single reference view, then selectively fills occluded or missing regions using additional references while preserving a coherent 3D structure even under large viewpoint changes (see Fig.~\ref{fig:intro}). 
The framework is built on three key insights:
\textit{i) Multi-View Consistency} - inpainting starts from a single reference view and propagates edits to additional views with controlled consistency checks;
\textit{ii) Depth-Aware Optimization} - Since monocular depth accuracy varies with distance from the image plane, we reweight its contribution during training to improve reliability; and
\textit{iii) Exploiting Instance Priors} - instance segmentation in training images enables the embedding of a 3D segmentation field in the Gaussian representation, ensuring that Gaussians align with object boundaries for more accurate reconstructions.

Following these principles, \modelname{} is designed to operate in seven stages: \emph{(i)} select a reference view and perform 2D inpainting of its masked region; \emph{(ii)} estimate monocular depth maps for both the inpainted reference and other masked views; \emph{(iii)} use depth maps only to group pixels by depth level, providing reliable depth priors; \emph{(iv)} incorporate object-based supervision by associating Gaussians with instance-level features, enabling the identification and isolation of individual objects for targeted editing; \emph{(v)} initialize and optimize Gaussians in missing regions using the inpainted image, depth priors, and object-aware constraints; \emph{(vi)} detect depth inconsistencies by comparing normalized gradients of monocular depth with rendered depth, then select the most inconsistent view as the new reference and apply 2D inpainting to its inconsistent region; \emph{(vii)} repeat this loop until all views meet geometric and visual consistency requirements.
This iterative strategy ensures consistent, high-quality 3D scene inpainting, leveraging both depth-aware and object-aware supervision to improve geometric coherence and enable controllable scene editing across all viewpoints.

Our contributions can be summarized as follows:

\begin{enumerate}[]
    \item We introduce a novel weighted depth optimization loss that reduces errors in monocular depth estimation, ensuring robust geometric continuity in the reconstructed 3D scenes. Unlike existing approaches, this loss does not require accounting for scale and shift alignment.
    \item We present a structured inpainting framework that systematically identifies inconsistencies among inpainted elements and the available training views. It then corrects these inconsistencies view by view while maintaining compatibility with previously processed views.
    \item We incorporate additional reconstruction priors to guide inpainting, encouraging Gaussians to align with object structures. This inherent grouping of Gaussians by objects benefits tasks such as 3D object selection.
   
\end{enumerate}

\begin{figure*}[htbp]
    \centering
    \includegraphics[width=\textwidth]{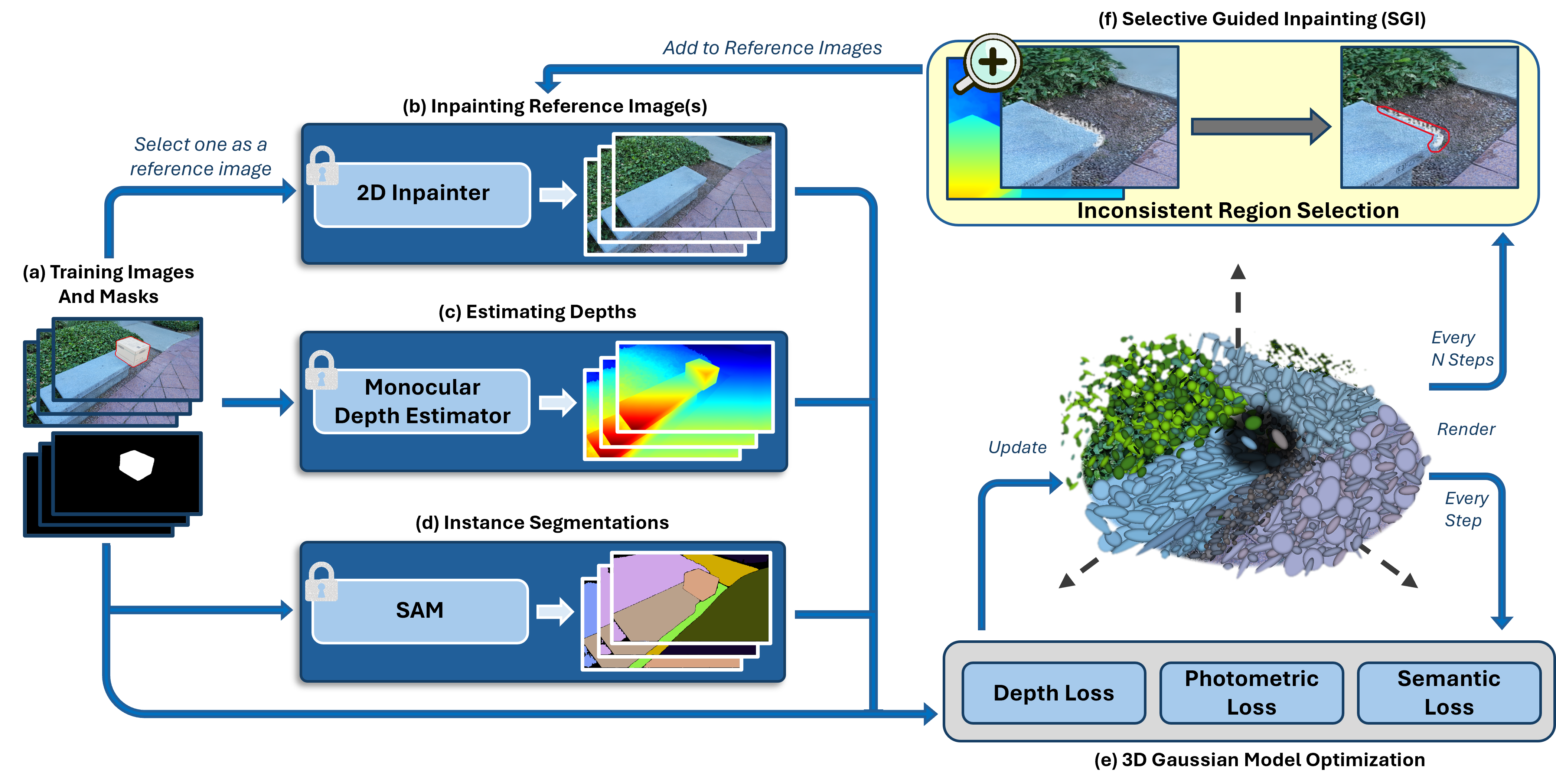}
    \caption{\textbf{Pipeline.} (a) We begin with a set of multi-view training images, each with its corresponding binary mask and camera pose. 
    (b) One image is selected as the reference and inpainted using a diffusion-based 2D inpainting model, providing plausible content for the masked region. 
    (c) Monocular depth estimation is performed for each image, producing depth maps that guide the 3D spatial placement of Gaussians.
    (d) Instance segmentation is applied to all images to generate object masks, providing semantic priors for the 3D Gaussians.
    (e) The 3DGS model is optimized with supervision from the photometric, depth, and object-aware signals, producing geometrically and semantically consistent inpainting across all views.
    (f) Selective Guided Inpainting (SGI) is iteratively applied by identifying and refining the regions with the highest inpainting errors using updated 2D inpainting on challenging views.}
    \label{fig:my_label}
\end{figure*}

\section{Related Work}
\label{sec:related}

In this section, we review prior work in image inpainting and its extensions to 3D scene reconstruction, focusing on three main areas: 2D inpainting methods, 3D scene inpainting approaches for NeRF and 3DGS models.

\subsection{2D inpainting}
\label{subsec:2d_inpainting}

Traditional 2D inpainting techniques initially relied on patch-based methods that leveraged repetitive patterns in images~\cite{bertalmio2000image, criminisi2004inpainting}. These methods found applications in interactive image editing~\cite{barnes2009patchmatch} and live video stream manipulation~\cite{herling2012pixmix}. Data-driven methods as convolutional neural networks~\cite{pathak2016context, iizuka2017globally, yu2019gatedconv} and generative adversarial networks~\cite{yu2018contextattention, zhao2021large} improved further the visual synthesis quality. More recently, diffusion models~\cite{rombach2022stablediffusion,Lugmayr2022repaint,xie2023smartbrush,Corneanu2024latentinpaint} and Fourier convolutions~\cite{Suvorov2022fourierconv} have achieved state-of-the-art performance in filling missing regions. While these methods excel in generating high-quality 2D images, they lack mechanisms for ensuring geometric consistency required in 3D scene inpainting. In practice, previous methods often apply an of-the-shelf 2D inpainting approach on one or more views to synthesize content in the missing regions, and then use these inpainted views as the starting point to perform 3D scene inpainting. We use the diffusion-based method from~\cite{esser2024scaling} to inpaint our reference views. 

\subsection{NeRF-based 3D scene inpainting methods}
\label{subsec:3d_inpainting_nerf}

The pionnering NeRF-based inpainting method in SPIn-NeRF \cite{mirzaei2023spin} inpaints each view independently and leverages a mean squared error loss on the unmasked regions and a perceptual loss on masked regions to guide NeRF optimization. In contrast, InNeRF360 \cite{wang2024innerf360} while also inpainting views independently, learns 3D shapes from 2D images using \cite{shapenet2015} to enforces global geometric consistency for full 360${^{\circ}}$ scenes. While these methods reduce blurriness, residual inconsistencies across views can still lead to geometric artifacts. To address these issues, more recent diffusion‐guided methods such as MVIP‐NeRF \cite{MVIPNeRF} and Inpaint3D \cite{prabhu2023inpaint3d}, directly incorporate diffusion priors via Score Distillation Sampling (SDS) to jointly optimize appearance and geometry. Another diffusion-based method NeRFiller \cite{weber2023nerfiller}, employs a tiling-based strategy to maintain geometric alignment across the inpainted regions. However, these approaches still often suffer from blurry details. Similarly, methods like OR–NeRF \cite{yin2023or} and MALDNeRF \cite{lin2024maldnerf}, which primarily rely on pixel–wise losses or architectural modifications, also tend to produce blurry outputs. An alternative, reference guided approach \cite{mirzaei2023reference} anchors the reconstruction to a single inpainted view, but frequently struggles to capture view‐dependent effects when target views deviate significantly from the reference. In contrast, while ours is not a NeRF-based inpainting method, it combines both single-reference and multi-reference strategies in an iterative framework, to achieve sharp, geometrically consistent reconstructions across diverse viewpoints, even in the presence of large occlusions. We compare our results with those presented in SPIn-NeRF \cite{mirzaei2023spin} and OR–NeRF \cite{yin2023or}.

\subsection{3DGS-based 3D scene inpainting methods}
\label{subsec:3d_inpainting_3dgs}
3DGS-based inpainting uses a sparse set of 3D Gaussians, each defined by explicit parameters like position, covariance, color, and opacity, to enable fast rendering and high-detail reconstruction. As opposed to NeRF’s implicit neural network representation, this clear, editable representation integrates easily into existing 3D pipelines.

InFusion \cite{liu2024infusion} starts by inpainting depth on a reference view, then unprojects the depth map to place 3D Gaussians in missing regions, which are fine-tuned to match the inpainted content. For complex scenes, it progressively uses multiple references. Our approach also works iteratively from a single reference to multi-reference views, but without explicit depth inpainting or training a latent diffusion model. Instead, we use depth cues from multiple views to guide 3DGS inpainting. GScream \cite{wang2025gscream} conditions scene completion on a single reference using depth cues from all views, but depth estimation errors and large viewpoint variations can introduce artifacts. In contrast, our method introduces a depth optimization loss that aligns estimated depth with a robust, structure-aware average along constant-depth contours, improving the capture of geometric structure and mitigating depth prior errors. RefFusion \cite{mirzaei2024reffusion} adapts a 2D inpainting diffusion model to a single reference and distills generative priors into a 3D scene using multi-view score distillation. However, its reliance on a single reference and auxiliary depth regularization limits its ability to handle complex scene variations, causing artifacts with imperfect references or depth estimates. Our method overcomes these limitations by using the depth optimization loss and a progressive strategy to incorporate additional views. We compare our results to \cite{liu2024infusion, mirzaei2024reffusion, wang2025gscream}, the closest to our method.

\section{Method}
\label{sec:method}

Given a set of \(N\) multi-view images \(\{I_i\}_{i=1}^N\) with their corresponding camera poses \(\{P_i\}_{i=1}^N\) and binary masks \(\{M_i\}_{i=1}^N\) that indicate observed and masked (to be inpainted) regions, our goal is to reconstruct a complete and consistent 3D scene via Gaussian Splatting, in which the inpainted regions exhibit geometric and perceptual fidelity from all viewpoints. Each 3D Gaussians are characterized by their 3D position \( \boldsymbol{\mu} \), a covariance matrix \( \Sigma \) that encapsulates scale and shape, color \( \mathbf{c} \), opacity \( \alpha \), and spherical harmonics coefficients to capture view-dependent appearance. These parameters are optimized such that rendering the model on one of the provided camera poses $P_i$ yields the corresponding image $I_i$. This is achieved by optimizing the parameters through a differentiable rendering process, in which the color of each pixel \( p \) is obtained by projecting all 3D Gaussians on the plane and $\alpha$-blending them according to their distance. This is summarized by the formula:

\begin{equation}
    C(p) = \sum_{i=1}^{|\mathcal{G}_p|} \mathbf{c}_i\, \alpha_i' \prod_{j=1}^{i-1} \left(1 - \alpha_j'\right),
    \label{eq:alpha_blend}
\end{equation}
where \( \alpha_i' \) denotes the effective opacity of the \( i \)-th Gaussian at pixel \( p \).

\subsection{Model Initialization}
As shown in Figure~\ref{fig:my_label} (a), we begin with a set of multi-view images with their binary masks indicating regions requiring inpainting. To fill the missing area, we select one image as a \emph{reference image} $r_0$ and apply a diffusion-based 2D inpainting model \cite{esser2024scaling} to its masked region (Figure~\ref{fig:my_label} (b)). This generates a plausible initial content for the occluded area, providing a concrete starting point for the inpainting process. The inpainted reference, together with the original masked images from other views, is used to optimize the photometric parameters of a 3DGS model. The unmasked areas are supervised by all images, while the inpainted region is optimized using the reference image only. Additionally, we also incorporate the loss functions described in the following subsections, the soft depth clustering loss and the crop-focused depth loss (section \ref{subsec:sdcl}) as well as the object-aware contrastive loss (section \ref{subsec:oacl}) to guide the learning process. This integrated approach ensures that, from the outset, the model benefits not only from the inpainted content but also from enhanced depth supervision and object-aware feature discrimination.

We empirically found that an initial training of 8,000 steps is sufficient to establish a robust baseline model, where the inpainted regions are plausibly initialized and the 3DGS representation captures the overall scene structure. However, due to limited supervision and viewpoint diversity, inconsistencies and artifacts may still persist, particularly in challenging regions. To address these remaining errors and achieve multi-view consistency, we apply our selective guided inpainting strategy (section~\ref{subsec:sgi}), which iteratively identifies and refines the most problematic areas based on their depth and appearance discrepancies.

\subsection{Depth-based Supervision}
\label{subsec:sdcl}
The first optimization loss, exemplified in Figure~\ref{fig:my_label} (c), employs monocular depth estimation to enhance the scene geometry in both inpainted and non-inpainted regions. The key insight of this loss is to enforce consistency between monocular depth maps obtained with an off-the shelf estimator (in our case, Depth Anything \cite{yang2024depth}) and the rendered depth from the Gaussian \(d(p)\) for each pixel $p$. The discrepancy between rendered and estimated depth serves as a supervisory signal to guide the correct spatial positioning of the 3D Gaussians.

This loss forces the Gaussians to adhere to the surfaces identified by monocular depth estimation and implicitly, by applying the loss to multiple views, to be consistent with the volumetric occupancy of the scene elements they represent. 
For the non-masked areas, this loss can make the overall model more geometrically consistent, and discourages floating or isolated Gaussians. Conversely, in the area to be inpainted this loss provides an additional source of supervision, since when considering a new image it allows to identify inconsistent, outlier elements. Given these different scopes, we define
two separate loss terms - called respectively Soft Depth Clustering Loss and  Crop-Focused Depth Loss -  and then combine them. 

\paragraph{Soft Depth Clustering Loss (SDCL)} is designed as an alternative to traditional metric depth losses, reducing the impact of scale-and-shift misalignment. The loss is computed over the full image by first partitioning the ground truth depth map into evenly spaced depth bins. For each bin, we generate a mask \(\mathcal{M}_k\). Within each mask, we calculate the mean depth \(\mu_k\) based on the rendered depths \(d(p)\). Since both \(d(p)\) and \(\mu_k\) come from the rendered depth, they have the same numerical scale and, therefore, can be compared directly without the need of any scale or shift alignment. SDCL consists of two main components:

\textbf{Per-Bin Loss:} For each depth bin \(k\), we compute the mean absolute error between the rendered depth \(d(p)\) and the bin’s mean depth \(\mu_k\) over all pixels \(p \in \mathcal{M}_k\): 

\begin{equation}
L_k = \frac{1}{|\mathcal{M}_k|} \sum_{p \in \mathcal{M}_k} \left| d(p) - \mu_k \right|.
\end{equation}

This term quantifies how well the rendered depths cluster around the mean within each bin.

\textbf{Weighted Averaging:} To aggregate the per-bin losses, we compute a weighted average where each bin’s weight \(w_k\) reflects its relative reliability. In our approach, bins corresponding to farther depth intervals—where errors tend to be larger—are assigned lower weights. This prioritizes the supervision in regions where depth predictions are more accurate. 

The overall soft depth clustering loss is then defined as

\begin{equation}
\mathcal{L}_{\text{SDCL}} = \frac{\sum_{k} w_k \cdot L_k}{\sum_{k} w_k} = \frac{\sum_{k} \frac{w_k }{|\mathcal{M}_k|}\sum_{p\in \mathcal{M}_k} \left| d(p) - \mu_k \right|}{\sum_{k} w_k}.
\end{equation}

\paragraph{Crop-Focused Depth Loss (CFDL)}encourages better depth estimation in the masked regions, especially in the reference inpainted images. Every 9 iterations, we crop a region around the inpainted area; this region is randomly expanded also to include part of the non-inpainted area and provide additional context. The cropped region is then processed using the same binning and mask creation strategy as the full-image SDCL. Within this crop, a Gaussian-like weighting scheme is applied to assign higher weights to pixels near the center of the inpainted region, providing focused supervision where it matters most. 

The overall depth loss is then computed as a weighted sum of the global and localized components:

\begin{equation}
\mathcal{L}_{\text{depth}} = \mathcal{L}_{\text{SDCL}} + \kappa \, \mathcal{L}_{\text{CFDL}},
\end{equation}
where \(\kappa\), the coefficient balancing the contributions of global scene geometry and localized inpainted detail, is empirically set to \(\kappa = 25\).

\subsection{Object-based Supervision} 
\label{subsec:oacl}

Beyond producing a visually complete scene, our framework is designed to enable object-level control within the reconstructed 3D representation. This is important because many real-world editing tasks, such as removing an object from a scene, require isolating individual objects in the 3D space. Once objects are explicitly identified, they can be removed, and our inpainting pipeline can be applied to seamlessly fill the resulting gaps in the scene.

To make this possible, we integrate an Object-Aware Contrastive Loss (OACL) that learns a segmentation-aware feature representation for each Gaussian. To do so, each 3D Gaussian is augmented with a 16-dimensional learnable parameter for capturing segmentation features as done in \cite{cgc}.

First, we generate 2D segmentation masks $m_p$ for each input image \(I \in \mathbb{R}^{H \times W}\) using SAM2 \cite{ravi2024sam2segmentimages}. The number of segments identified by SAM2 in the image will be indicated as $N_k$. 
For each training image, we render a feature map of size ${H \times W \times 16}$ from a chosen viewpoint by splatting the Gaussians and alpha blending associated the 16-dimensional learnable parameter (similar to what is done for the color parameter $\mathbf{c}_i$ in Eq. (\ref{eq:alpha_blend})). We then group the rendered features based on their corresponding segmentation masks, forming clusters \(\{f_p\}\) for each segment \(m_p\). The centroid \(\bar{f}_p\) of each cluster is computed as the mean feature vector of that group. Our goal is to maximize the similarity of features within each cluster while minimizing the similarity between features across different clusters. This is achieved by minimizing the following loss function:

\begin{equation}
\mathcal{L}_{\text{OACL}} = -\frac{1}{N_k} \sum_{p=1}^{N_k} \frac{1}{|\{f_p\}|} \sum_{q=1}^{|\{f_p\}|} \log \frac{\exp\left( \frac{f_p^q \cdot \bar{f}_p}{\phi_p} \right)}{\sum_{s=1}^{N_k} \exp\left( \frac{f_p^q \cdot \bar{f}_s}{\phi_s} \right)},
\end{equation}
where \(f_p^q\) denotes the \(q\)-th feature in cluster \(\{f_p\}\) and \(\phi_p\) is the temperature parameter for the \(p\)-th cluster. We determine \(\phi_p\) based on the dispersion of features within the cluster:
\begin{equation}
\phi_p = \frac{1}{N_p} \sum_{q=1}^{N_p} \|f_p^q - \bar{f}_p\|_2 \cdot \log (N_p + \epsilon),
\end{equation}
with \(N_p = |\{f_p\}|\) and a stabilizing constant \(\epsilon = 100\). To ensure stable similarity measurements, all features are \(\ell_2\)-normalized prior to loss computation.

By incorporating this object-aware contrastive loss, our model learns robust and distinct object representations in the 3D feature space. This not only enhances the semantic coherence of the inpainted regions but also bridges the gap between static reconstruction and fully controllable 3D content creation.

\subsection{Selective Guided Inpainting}
\label{subsec:sgi}

Even with depth-based supervision and an initial inpainted reference view, achieving perfect multi-view consistency remains challenging. In practice, certain viewpoints, particularly those far from the initial reference, tend to exhibit incomplete geometry, blurred results, or inconsistencies in occluded regions. These errors often arise because the inpainting guidance from a single reference view does not fully constrain the geometry in occluded areas.

To address this, we introduce \emph{Selective Guided Inpainting} (SGI), an iterative refinement process that progressively identifies and corrects the most problematic regions in the reconstruction. The central idea is to avoid re-inpainting the entire masked regions for each training image, an approach that is computationally expensive and risks introducing unnecessary changes to regions that are already consistent, and instead focus only on views and sub-regions where inconsistencies are most severe.

At each refinement step, we render all training views from the current 3DGS model and compare their rendered depth maps \(D_{\text{rend}}(x,y)\) with monocular depth predictions \(D_{\text{mono}}(x,y)\) obtained from an off-the-shelf estimator. This comparison is restricted to the masked regions that require inpainting. We then compute an absolute depth error map
\begin{equation}
E(x,y) = \left| D_{\text{rend}}(x,y) - D_{\text{mono}}(x,y) \right|,
\end{equation}
which highlights areas where the current reconstruction deviates significantly from the geometric prior. The view \(v^*\) with the largest cumulative depth error \(\sum_{x,y} E(x,y)\) is selected for targeted refinement. Within the selected view \(v^*\), we further localize problematic areas by analyzing the spatial variation of the error map. Specifically, we compute the gradient magnitude
\begin{equation}
G(x,y) = \left\| \nabla E(x,y) \right\|_2,
\end{equation}
to detect sharp transitions indicative of geometric inconsistencies, such as misplaced surfaces or depth discontinuities. We then threshold \(G(x,y)\) to obtain a binary map, and apply morphological dilation to ensure full coverage of the inconsistent regions. The result is a refined binary mask \(B(x,y) \in \{0,1\}\) that marks the subset of pixels requiring re-inpainting.

Rather than modifying the entire view, we apply a diffusion-based 2D inpainting model only to the regions indicated by \(B(x,y)\). This targeted update strategy minimizes disruption to well-reconstructed areas while injecting high-quality, view-specific guidance exactly where it is needed. The newly inpainted image from iteration $k$, denoted $I_{\text{inp}}^{(k)}$, together with its updated mask, is then added to the set of reference views, providing additional supervision for subsequent optimization. 
Summing up, SGI first selects the view with the largest overall depth error, then localizes the geometric inconsistencies within it, and re-inpaints only those regions. The resulting updated image is added to the reference set, providing new guidance for subsequent optimization. By iteratively applying this refinement, the model is progressively completed, with each iteration addressing smaller and smaller discrepancies while preserving the appearance learned in previous steps.

\section{Experiments}
\label{sec:exp}

In this section, we evaluate the \modelname{} framework and compare it against state-of-the-art 3D inpainting methods. We first introduce the experimental setup (Sec.~\ref{sec:exp:setup}), then provide a detailed quantitative and qualitative evaluation of our approach (Sec.~\ref{sec:exp:experiments}) along with the training times. Finally, we provide a set of ablation experiments to assess the contribution of each of the three losses; the advantage of using our monocular depth loss over the existing implementation proposed by GScream (Sec.~\ref{sec:exp:ablation}); and the benefits of including our method to the original MCMC~\cite{kheradmand2024mcmc} approach.

\subsection{Experimental Setup}
\label{sec:exp:setup}

\paragraph{Dataset.} We evaluated our method using the SPIn-NeRF~\cite{mirzaei2023spin} dataset, a benchmark specifically prepared for object removal evaluation in forward-facing, in-the-wild scenes. The dataset comprises 10 scenes, each containing 100 multi-view images with human-annotated object masks. For each scene, 60 images include an unwanted object (used as training views), while the remaining 40 images capture the scene without the object (used as test views) and allow assessing the quality of the inpainted model.

%
%


\paragraph{Evaluation Metrics.} To evaluate the performance we employ four widely recognized metrics: Peak Signal-to-Noise Ratio (PSNR), Structural Similarity Index Measure (SSIM)~\cite{wang2004image}, Learned Perceptual Image Patch Similarity (LPIPS)~\cite{zhang2018unreasonable}, and Fréchet Inception Distance (FID)~\cite{heusel2017gans}. These metrics are computed both across the full image and within the specific object mask region to provide a comprehensive assessment of image quality and fidelity in both global and localized contexts. For the evaluations we indicate the best and the 2nd best values using \textbf{bold} and \underline{underlined}, respectively.

\paragraph{Implementation Details}  
The models were trained on a system equipped with NVIDIA A100 80G GPU and an Intel Xeon Silver 4316 CPU with 80 cores. Our implementation builds upon a modified version of the MCMC framework~\cite{kheradmand2024mcmc}, where we extended the CUDA rasterization functions to additionally render depth information.



\paragraph{Baselines}
We compare our method against four baseline approaches: GScream~\cite{wang2025gscream}, SPIn-NeRF~\cite{mirzaei2023spin}, and OR-NeRF~\cite{yin2023or}. 
To ensure a fair comparison with GScream, we retrain and evaluate the models using their publicly available open-source code. For the remaining methods, we utilize the results reported by GScream. Additionally, since GScream's evaluations were conducted on a slower GPU than ours, we provide appropriately scaled training times for a more accurate comparison with SPIn-NeRF and OR-NeRF.


\subsection{Experimental results}
\label{sec:exp:experiments}

\begin{figure*}[t!]
  \centering
  \includegraphics[width=\linewidth]{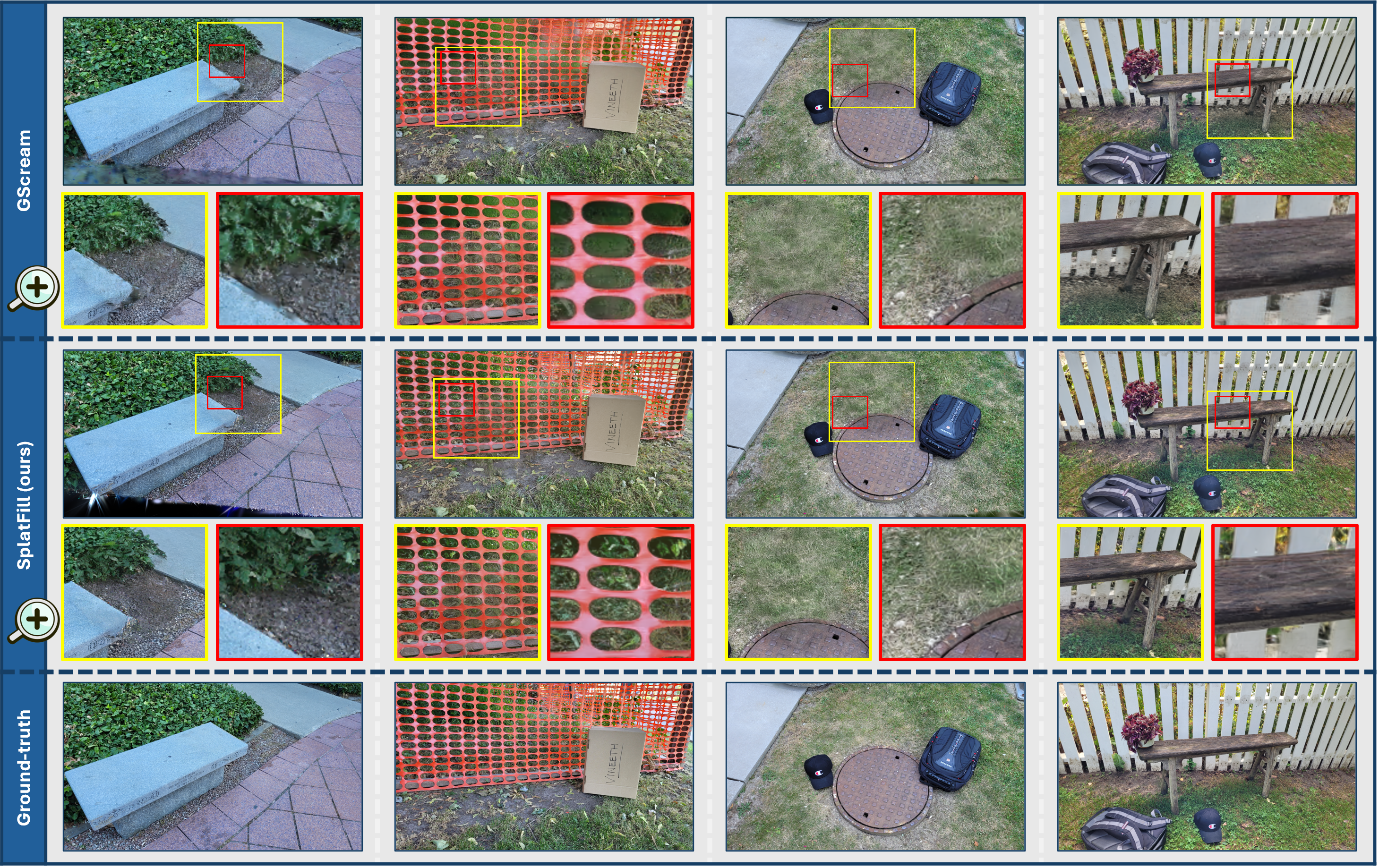}
  \caption{
    \label{fig:qualitative_basic}
    \textbf{Qualitative evaluation} across multiple scenes (each column shows a different scene). 
    For each method, the first row shows the full inpainted view, and the second row provides zoomed-in crops of selected regions from that view. 
    The top block corresponds to GScream, the middle block to our proposed method, and the bottom block to the ground truth, where the target objects were physically removed from the real environment.
  }
\end{figure*}

We evaluated our method across all scenes in the SPIn-NeRF dataset, reporting both qualitative and quantitative results. 

\paragraph{Quantitative Comparison}

We first report the average PSNR, LPIPS, and SSIM scores for the entire views in  Tab.~\ref{table:FullRegion}. 
The results in this table show that our method consistently outperforms SPIn-NeRF and OR-NeRF across all metrics. Against GScream, it delivers comparable performance in PSNR, SSIM, and LPIPS, but achieves a significantly lower FID score, reflecting a stronger alignment and coherence with the surrounding scene content.
Moreover, our method is also more efficient to train, with $\mathbf{\sim24.5\%}$\textbf{ faster training time} than GScream, the fastest baseline. 

In Tab.~\ref{table:MaskedRegion}, we present the results when considering only the masked regions. Our method performs on par with GScream across all metrics.  While SPIn-NeRF and OR-NeRF show higher reported scores, these values are taken directly from GScream’s results and may not fully reflect differences in conditions. Given that GScream has already been shown to outperform both SPIn-NeRF and OR-NeRF on the same dataset and evaluation setup, our comparable performance to GScream strongly indicates that our method also surpasses these baselines in the masked regions.

\begin{table}[ht]
\centering
\begin{adjustbox} {max width=\columnwidth}
\begin{tabular}{|l|c|c|c|c|c|}
\hline
\textbf{Methods} & \textbf{PSNR ↑} & \textbf{SSIM ↑} & \textbf{LPIPS ↓} & \textbf{FID ↓} & \textbf{Training Time ↓} \\ \hline
SPIn-NeRF        & 20.18*          & 0.46*          & 0.47*          & 58.78*       & $\sim$ 112.5 mins*           \\ \hline
OR-NeRF          & 20.32*          & 0.54*          & 0.35*          & 38.69*       & $\sim$ 225*                  \\ \hline
GScream          & \underline{20.45}           & \underline{0.58}           & \underline{0.28}           & \underline{36.72}         & \underline{45 mins}                   \\ \hline
Ours             & \textbf{20.46}           & \textbf{0.63}           & \textbf{0.25 }          & \textbf{29.76}         & \textbf{34 mins}                   \\ \hline
\end{tabular}
\end{adjustbox}
\caption{Average results for all scenes in the SPIn-NeRF dataset (Full Region). `*' indicates the metrics are based on those reported in GScream.}
\label{table:FullRegion}
\end{table}
\begin{table}[h]
\centering
\begin{adjustbox} {width=0.7\columnwidth}
\begin{tabular}{|l|c|c|c|}
\hline
\textbf{Methods} & \textbf{PSNR ↑} & \textbf{SSIM ↑} & \textbf{LPIPS ↓} \\ \hline
SPIn-NeRF        & \textbf{15.80*}          & \textbf{0.21*}          & 0.58*          \\ \hline
OR-NeRF          & \underline{15.74*}          & \textbf{0.21*}          & \underline{0.56*}          \\ \hline
GScream          & 15.67           & \textbf{0.21}           & \textbf{0.54}           \\ \hline
Ours             & 15.67           &  \textbf{0.21}          & \textbf{0.54}          \\ \hline
\end{tabular}
\end{adjustbox}
\small
\caption{Average results for all scenes in the SPIn-NeRF dataset (Masked Region). `*' indicates the metrics are based on those reported in GScream.}
\label{table:MaskedRegion}
\end{table}

\paragraph{Qualitative Comparison.}
The improvements observed in our quantitative evaluation are further confirmed by the qualitative results shown in Figure~\ref{fig:qualitative_basic}. Overall, our method produces inpainted regions with sharper textures and improved geometric coherence compared to GScream. 
In the first column, our method reconstructs the stone bench with sharper edges and a more natural integration into the surroundings, while GScream exhibits noticeable inconsistencies along the right edge of the bench. 
In the second column, the grass behind the fence is reconstructed with significantly finer detail and smoother blending in our result, whereas GScream’s output appears softer and less coherent. 
The third column highlights our ability to recover detailed grass patterns around the manhole, preserving the natural texture that is partially lost in GScream’s inpainting. 
Finally, in the fourth column, the distant grass beyond the fence remains sharp and structurally consistent, while GScream’s result becomes noticeably blurred in background areas. The zoomed-in crops in Figure~\ref{fig:qualitative_basic} clearly illustrate these differences, providing visual evidence that our method better preserves structural detail and spatial consistency in the inpainted regions.

\subsection{Ablation Study}
\label{sec:exp:ablation}

While the previous section establishes the overall effectiveness of our method, here we analyze the contributions of its main components to better understand their impact on 3D inpainting.

\paragraph{Depth-based supervision.}  
We first examine the role of our depth loss by comparing three configurations: (i) no depth supervision, (ii) the depth loss formulation used in GScream, and (iii) our full depth-based supervision combining the Soft Depth Clustering Loss (SDCL) and Crop-Focused Depth Loss (CFDL).  
The results in Table~\ref{table:depth_ablation} show a clear advantage when using our full depth loss: both SDCL and CFDL contribute to better spatial placement of Gaussians, which in turn reduces geometric inconsistencies and improves the perceptual quality of the inpainted regions. Qualitative comparisons and further discussion are provided in the supplementary material.

\paragraph{Inpainting strategy.}  
We next evaluate the impact of our Selective Guided Inpainting (SGI) approach. For this, we compare (i) training with a single inpainted reference image, and (ii) our full SGI pipeline that iteratively identifies the most inconsistent views, refines them locally, and incorporates them back into the reference set.  
As reported in Table~\ref{table:sgi_ablation}, SGI improves performance over the single-reference setup. Qualitative examples illustrating this improvement are included in the supplementary material.

Overall, these ablations confirm that both our depth-based supervision and our SGI strategy are essential for the robustness and quality of our 3D inpainting results: the depth loss anchors geometry in both visible and occluded areas, while SGI ensures that remaining inconsistencies are removed in a targeted and stable manner.

\begin{table}[h]
\centering
\begin{adjustbox} {max width=\columnwidth}
\begin{tabular}{|l|c|c|c|}
\hline
\textbf{Depth-based supervision} & \textbf{PSNR ↑} & \textbf{SSIM ↑} & \textbf{LPIPS ↓} \\ \hline
None           & 15.14 & 0.18 & \underline{0.56} \\ \hline
GScream Depth Loss     & \underline{15.22} & \underline{0.19} & \underline{0.56} \\ \hline
Our Full Depth Loss        & \textbf{15.67} & \textbf{0.21} & \textbf{0.54} \\ \hline
\end{tabular}%
\end{adjustbox}
\small
\caption{Ablation on depth-based supervision: Comparison among our full depth loss (SDCL + CFDL), no depth loss, and the GScream depth loss variant. (masked region)}
\label{table:depth_ablation}
\end{table}

\begin{table}[h]
\centering
\begin{adjustbox} {max width=\columnwidth}
\begin{tabular}{|l|c|c|c|}
\hline
\textbf{Inpainting strategy} & \textbf{PSNR ↑} & \textbf{SSIM ↑} & \textbf{LPIPS ↓} \\ \hline
Single Reference             & \underline{15.35} & \underline{0.19} & \textbf{0.54} \\ \hline
Selective Guided Inpainting (SGI) & \textbf{15.67} & \textbf{0.21} & \textbf{0.54}  \\ \hline
\end{tabular}%
\end{adjustbox}
\small
\caption{Ablation on the inpainting strategy: Comparison among using a single reference view and our proposed Selective Guided Inpainting (SGI) approach. (on masked region)}
\label{table:sgi_ablation}
\end{table}



\section{Limitations}
While our method improves both quality and training speed compared to existing approaches, several challenges remain. The quality of the initial reference view has a strong influence on how close the initialized model is to the desired reconstruction; in this work, we select it randomly, but more informed heuristics could yield better results. Our use of segmentation is also limited: while it clusters Gaussians by object, it is not yet fully exploited for inpainting, as we do not segment and complete each 3D object independently. Finally, the current approach does not explicitly handle illumination changes between views, meaning that variations such as shadows or reflections can lead to inconsistencies.

\section{Conclusion}
\label{sec:conc}

\modelname{} introduces a 3D scene inpainting framework for Gaussian Splatting with depth-guided and object-aware supervision, along with a consistency-aware refinement strategy. Our approach enhances coherence across views while achieving a 24.5\% faster training time compared to existing methods. Experiments on SPIn-NeRF show that \modelname{} achieves state-of-the-art performance, offering a consistent improvement across all metrics. Additionally, it reduces blurriness and artifacts, particularly in challenging viewpoint changes, resulting in sharper inpainted details and better perceptual quality. This makes \modelname{} a computationally efficient and visually robust strategy for 3D scene inpainting. Future work could further refine illumination consistency and reference view selection to further enhance 3D scene inpainting quality.

{
    \small
    \bibliographystyle{ieeenat_fullname}
    \bibliography{main}
}

\clearpage
\setcounter{page}{1}
\maketitlesupplementary

\section{Object Insertion}

\begin{figure}[h]
  \centering
  \includegraphics[width=\linewidth]{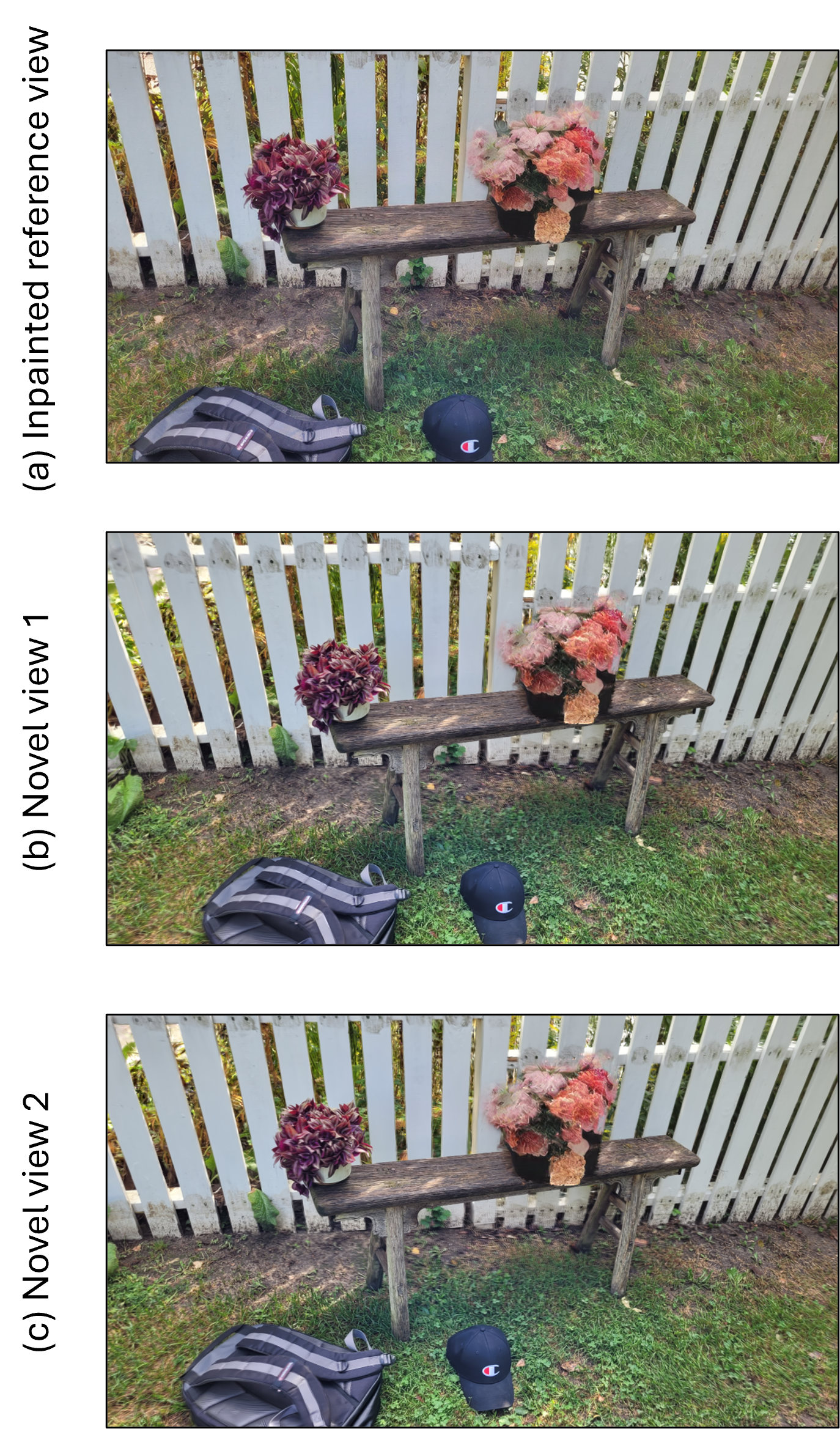}
  \caption{Object insertion results. (a) Inpainted reference view generated by Stable Diffusion using the prompt “flowers on the bench”. (b) and (c) are two novel views produced by our model.}
  \label{fig:object_insertion}
\end{figure}

In addition to 3D inpainting, our framework is capable of seamlessly inserting new objects into the reconstructed 3D scene. For object insertion, we adopt a strategy analogous to our inpainting approach. Specifically, the user provides a textual prompt to Stable Diffusion (e.g., “flowers on the bench”) to generate a reference view with the desired object inserted into the scene. This 2D inpainting output serves as the reference image for our pipeline. Following the same selective guided inpainting strategy used for missing regions, the inpainted reference is then incorporated into our 3D Gaussian Splatting model, ensuring that the inserted object is rendered with high-frequency details and consistent spatial alignment across different viewpoints. Figure~\ref{fig:object_insertion} illustrates this process: (a) shows the inpainted reference view produced by Stable Diffusion, while (b) and (c) present two novel views generated by our model.

\section{Additional Ablation}

\begin{figure*}[h]
  \centering
  \includegraphics[width=\textwidth]{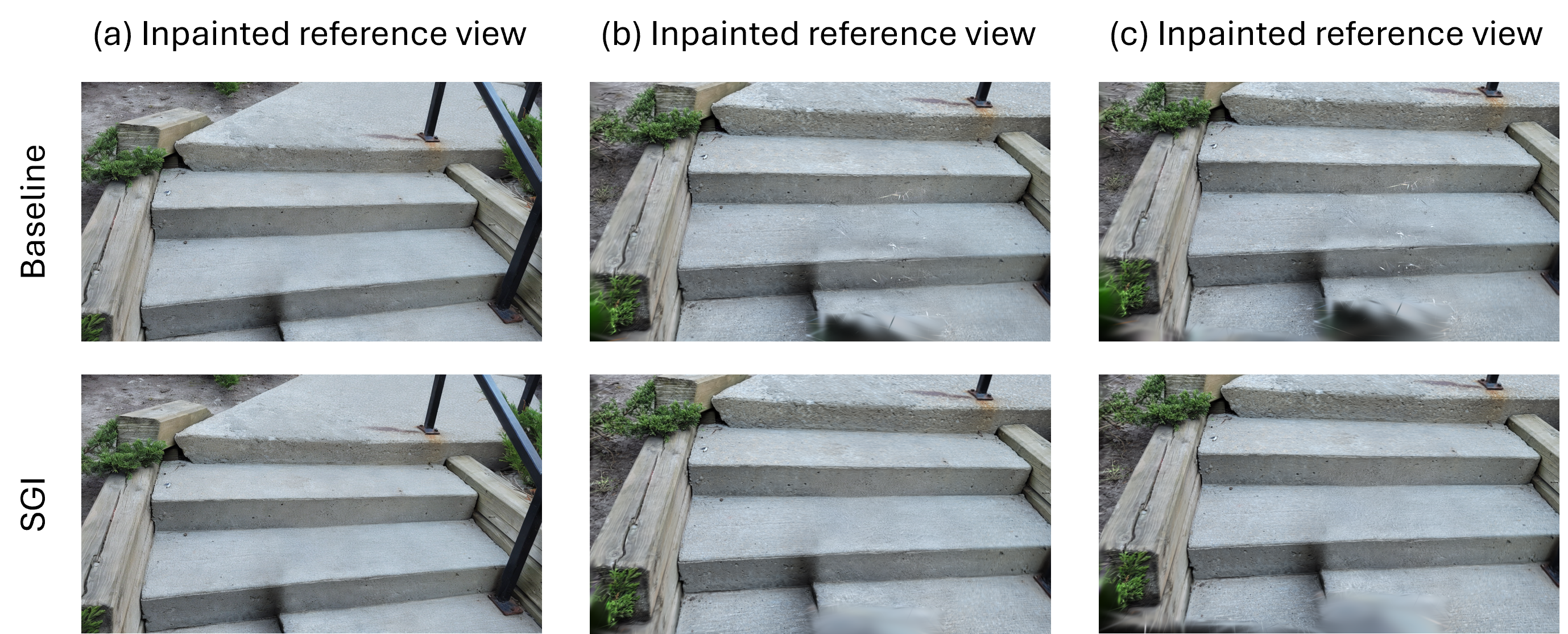}
  \caption{Comparison of inpainting results spanning two columns. \textbf{Top row (Baseline: Reference-only):} The inpainted reference view is used as guidance for rendering novel views, resulting in blurred and inconsistent details when the viewpoint diverges significantly from the reference. \textbf{Bottom row (Selective Guided Inpainting - SGI):} Our approach selectively refines regions with high inpainting error, yielding sharper and more coherent novel views. In both rows, the left column shows the inpainted reference view, while the middle and right columns display two novel views rendered from distant viewpoints.}
  \label{fig:sgi_comparison}
\end{figure*}

\paragraph{Effect of Selective Guided Inpainting}
To evaluate the impact of our Selective Guided Inpainting (SGI) strategy, we compare two methods: one that relies solely on the inpainted reference view as guidance and our full SGI approach. Figure~\ref{fig:sgi_comparison} is organized into two rows corresponding to these methods, with three columns in each row. In the left column, the inpainted reference image generated by the 2D inpainting model is shown; the middle and right columns display two novel views rendered from viewpoints that diverge significantly from the reference. The baseline method (top row) suffers from inconsistent details in the novel views due to insufficient guidance when far from the reference, whereas our SGI approach (bottom row) selectively refines regions with high inpainting error, yielding more coherent reconstructions across diverse viewpoints. 

\paragraph{Effect of Depth Loss Supervision}

\begin{figure*}[h]
  \centering
  \includegraphics[width=\textwidth]{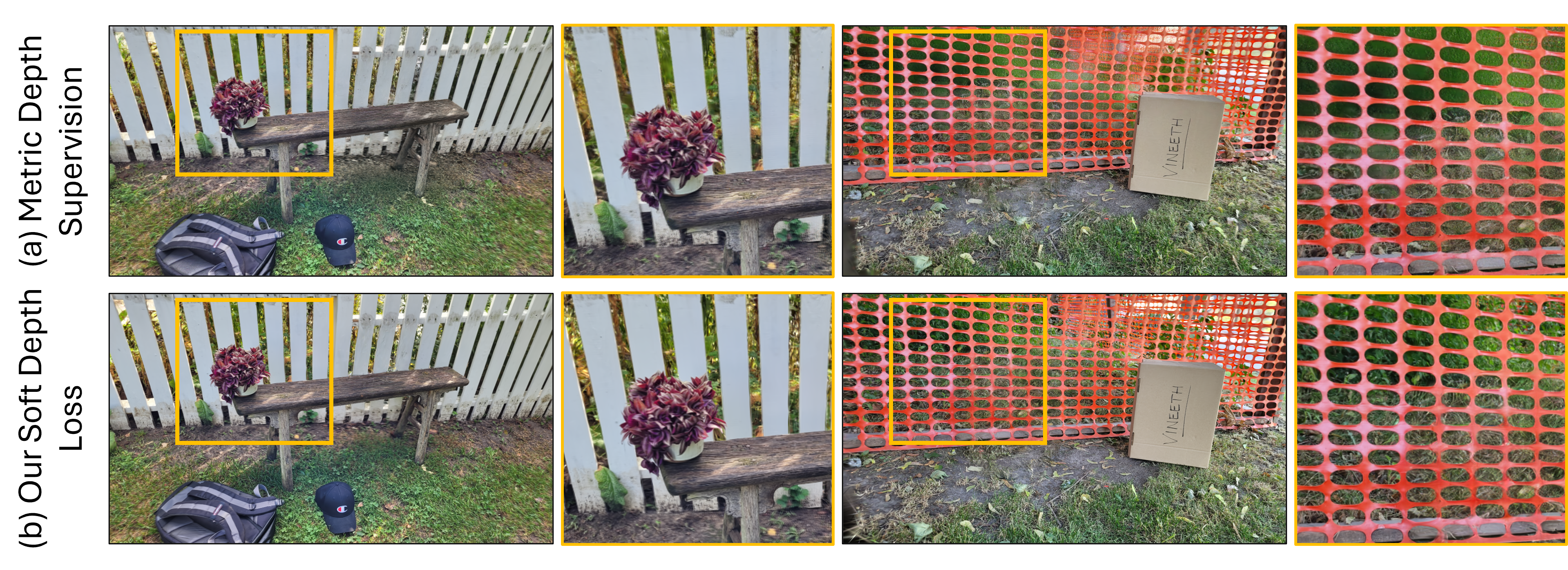}
  \caption{Qualitative comparison of depth supervision strategies. \textbf{Top row (a):} Reconstruction using conventional metric depth supervision, where distant objects appear blurred due to scale-and-shift alignment errors. \textbf{Bottom row (b):} Reconstruction using our soft depth loss, which yields sharper and more detailed reconstructions, even for objects at far distances.}
  \label{fig:depth_loss_comparison}
\end{figure*}

To further evaluate the impact of our depth loss formulation, we compare qualitative results from models trained with different depth supervision strategies. In Figure~\ref{fig:depth_loss_comparison}, the top row (a) shows the reconstruction results obtained using conventional metric depth supervision, where errors in scale-and-shift alignment lead to noticeable blurring of distant objects. In contrast, the bottom row (b) illustrates the results using our soft depth loss, which produces more detailed reconstructions even for far objects. This comparison demonstrates that our soft depth loss effectively mitigates alignment errors and captures fine geometric details across varying depths.

\begin{figure*}[h]
  \centering
  \includegraphics[width=0.85\textwidth]{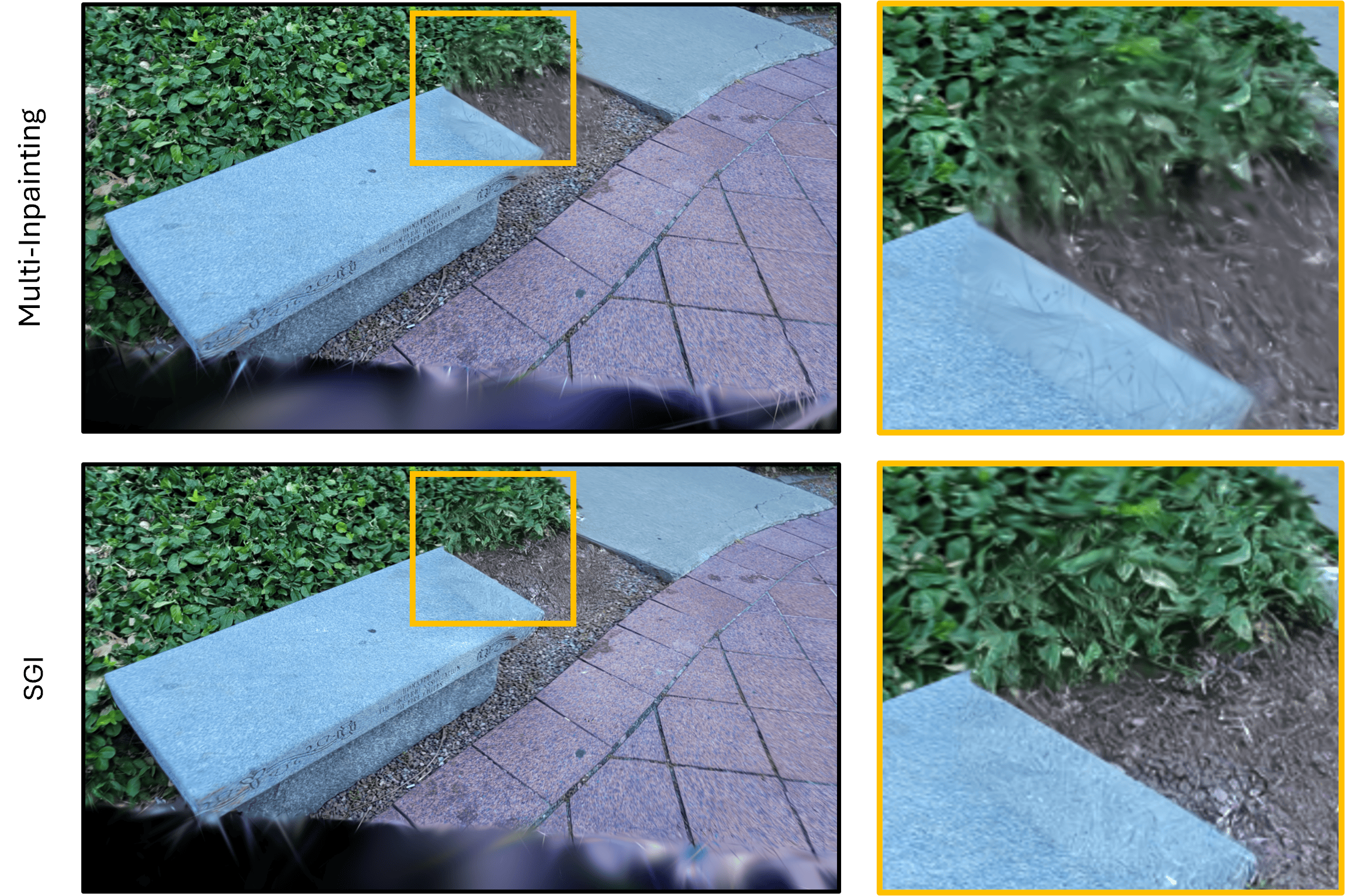}
  \caption{Qualitative comparison of multi-inpainting strategies. \textbf{Top row (Multi-Inpainting Baseline):} Uniform inpainting across all views leads to oversmoothed and blurred results. \textbf{Bottom row (Selective Guided Inpainting - SGI):} Our method selectively refines regions with high error, resulting in sharper and more detailed inpainted outputs.}
  \label{fig:multi_inpainting_ablation}
\end{figure*}

\paragraph{Multi-Inpainting Without Selective Mask}
To further validate the effectiveness of our SGI strategy, we compare our approach against a baseline that performs multi-view 2D inpainting without selective refinement. In the baseline, multiple views are inpainted without isolating the regions exhibiting high error, which leads to oversmoothed and blurred outputs. Figure~\ref{fig:multi_inpainting_ablation} presents a qualitative comparison: the top row shows the results of the baseline multi-inpainting approach, whereas the bottom row displays our Selective Guided Inpainting (SGI) method. The SGI strategy selectively refines only the regions with significant errors, yielding sharper and more detailed inpainted outputs.

\section{Additional Implementation Details}
In this section, we provide further details regarding the implementation of our \modelname{} framework.
Our implementation builds upon a modified version of the MCMC framework~\cite{kheradmand2024mcmc}, where we extended the CUDA rasterization functions to render depth information alongside color. This modification allows us to compute the rendered depth \(d(p)\) for each pixel \(p\) and to incorporate our soft depth clustering loss (SDCL) and crop-focused depth loss (CFDL) during training. Additionally, we modified MCMC to augment each 3D Gaussian with a 16-dimensional learnable p dedicated to capturing segmentation features. Training was performed on a system equipped with an NVIDIA A100 80G GPU and an Intel Xeon Silver 4316 CPU with 80 cores.

A total of 8000 training steps were used to establish a robust baseline before further refinement with our Selective Guided Inpainting (SGI) strategy. The following hyperparameters were employed:
\begin{itemize}
    \item \textbf{Soft Depth Clustering Loss (SDCL):} Computed at every 59 training iterations.
    \item \textbf{Crop-Focused Depth Loss (CFDL):} Computed every 9 iterations.
    \item \textbf{Weighting factor:} The CFDL contribution is balanced by a factor of \(\kappa = 25\).

\end{itemize}

Our Selective Guided Inpainting (SGI) module employs an error-driven approach to identify regions that require further refinement. Specifically, for each training camera view, we compute an error heatmap by measuring the per-pixel absolute difference between the rendered depth from the 3DGS model and the ground-truth depth estimated via a monocular depth estimator, restricted to the masked regions. The module processes all training views by first rendering the scene and saving the corresponding images and depth maps. To ensure comparability between the rendered and estimated depth maps - despite potential scale differences, both are computed using consistent camera parameters and normalized to a common metric scale during pre-processing. For each view (excluding the reference), it computes a cumulative depth error over the masked area and selects the view with the highest error as the candidate for further refinement. Once the candidate view is identified, the function analyzes its error map by computing the gradient magnitude using the Sobel operator, which reveals abrupt changes in depth error. To robustly detect these regions, the base mask is first eroded to exclude boundary artifacts, and the gradient map is thresholded to generate a binary mask highlighting regions with significant discrepancies. This binary mask is further refined via morphological dilation to ensure that all problematic areas are adequately covered. 

\end{document}